\documentclass{article}

    \PassOptionsToPackage{numbers, compress}{natbib}

\usepackage[final]{neurips_2025}

\usepackage[utf8]{inputenc} 
\usepackage[T1]{fontenc}    
\usepackage{hyperref}       
\usepackage{url}            
\usepackage{booktabs}       
\usepackage{amsfonts}       
\usepackage{nicefrac}       
\usepackage{microtype}      
\usepackage{xcolor}         

\usepackage{multirow}
\usepackage{graphicx}
\usepackage[table,xcdraw]{xcolor}
\usepackage{amsmath}
\usepackage{cleveref}
\usepackage{caption}

\title{Optimized Minimal 3D Gaussian Splatting}

\author{%
  Joo Chan Lee \\
  Sungkyunkwan University\\
  Suwon, South Korea \\
  \texttt{maincold2@skku.edu} \\
  \And
  Jong Hwan Ko\footnotemark[1]  \\
  Sungkyunkwan University \\
  Suwon, South Korea \\
  \texttt{jhko@skku.edu} \\
  \And
  Eunbyung Park\footnotemark[1]  \\
  Yonsei University \\
  Seoul, South Korea \\
  \texttt{epark@yonsei.ac.kr} \\
}

\begin{document}
{%
\renewcommand{\thefootnote}{*}
\footnotetext{Corresponding authors}
\renewcommand{\thefootnote}{\arabic{footnote}}
\maketitle
\begin{center}
    \centering
    \captionsetup{type=figure}
    \includegraphics[width=1.0\linewidth]{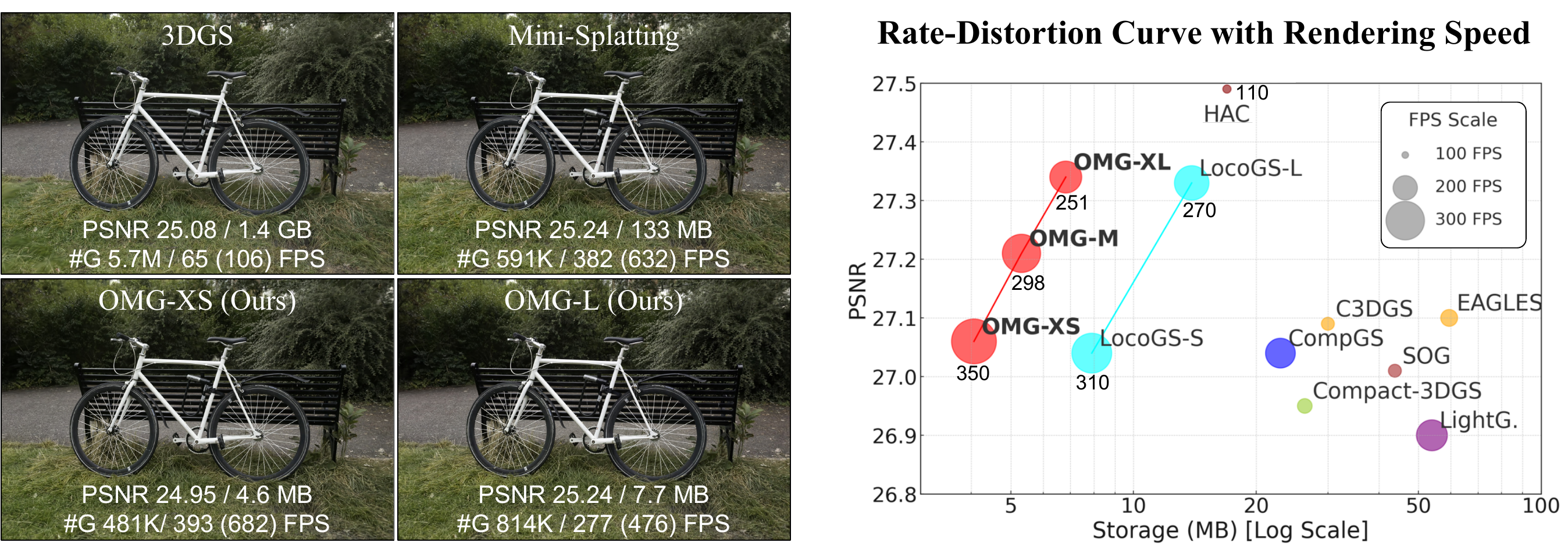}
    \captionof{figure}{Our approach focuses on minimizing storage requirements while using only a minimal number of Gaussian primitives. To achieve this, we introduce a local distinctiveness metric to identify the important Gaussians. Additionally, we propose a more efficient attribute representation, particularly for sparse Gaussians, by exploiting their irregularity and continuity. As a result, our method enables scene representation under 5 MB while achieving 600+ FPS rendering.}
    \label{fig:demo}
\end{center}%
}

\begin{abstract}
3D Gaussian Splatting (3DGS) has emerged as a powerful representation for real-time, high-performance rendering, enabling a wide range of applications. However, representing 3D scenes with numerous explicit Gaussian primitives imposes significant storage and memory overhead. Recent studies have shown that high-quality rendering can be achieved with a substantially reduced number of Gaussians when represented with high-precision attributes. Nevertheless, existing 3DGS compression methods still rely on a relatively large number of Gaussians, focusing primarily on attribute compression. This is because a smaller set of Gaussians becomes increasingly sensitive to lossy attribute compression, leading to severe quality degradation. Since the number of Gaussians is directly tied to computational costs, it is essential to reduce the number of Gaussians effectively rather than only optimizing storage.
In this paper, we propose Optimized Minimal Gaussians representation (OMG), which significantly reduces storage while using a minimal number of primitives. First, we determine the distinct Gaussian from the near ones, minimizing redundancy without sacrificing quality. Second, we propose a compact and precise attribute representation that efficiently captures both continuity and irregularity among primitives. Additionally, we propose a sub-vector quantization technique for improved irregularity representation, maintaining fast training with a negligible codebook size.
Extensive experiments demonstrate that OMG reduces storage requirements by nearly 50\% compared to the previous state-of-the-art and enables 600+ FPS rendering while maintaining high rendering quality.
Our source code is available at \href{https://maincold2.github.io/omg/}{https://maincold2.github.io/omg/}.
\end{abstract}

\section{Introduction}
3D Gaussian Splatting (3DGS)~\cite{3dgs} has gained popularity for fast and photorealistic 3D scene reconstruction and rendering, offering a compelling alternative to conventional methods. By leveraging tile-based parallelism to approximate NeRF’s~\cite{nerf} volumetric rendering, 3DGS enables significantly accelerated rendering while maintaining high visual quality. This has facilitated a wide range of applications, such as dynamic scene reconstruction~\cite{realtime4dgs, shapeofmotion}, photorealistic avatar generation~\cite{3dgsavatar, human3dgs}, generative models~\cite{dreamgaussian, text3dgs}, and city-scale rendering~\cite{hierarch3dgs, octreegs}, demonstrating its versatility across various domains.

3DGS adjusts the number of Gaussian primitives during training by iteratively cloning or splitting Gaussians with high positional gradients while removing low-opacity Gaussians. However, this optimization process introduces a substantial number of redundant Gaussians (over 3 million per 360 scenes~\cite{mip360}), leading to excessive storage requirements and computational overhead. To address this issue, various approaches have been proposed, including pruning based on rendering loss~\cite{c3dgs, maskgaussian} or importance score~\cite{radsplat, msv1} and optimized densification strategies~\cite{taming3dgs}. Notably, several methods~\cite{msv1, msv2, gaussianspa} reduce the number of Gaussians to around 0.5 million, enabling real-time rendering even on low-capacity GPUs while preserving rendering quality.

Despite these efforts, reducing the number of Gaussians alone does not sufficiently mitigate storage overhead. Each Gaussian is parameterized by 59 learnable parameters, so even with a reduced number of primitives, storage consumption remains substantial (e.g., 133 MB in \Cref{fig:demo}). To address this, many works have explored compressing Gaussian attributes by leveraging vector quantization~\cite{compact3d, lightgaussian}, neural fields~\cite{locogs}, sorting mechanisms~\cite{c3d2d}, and entropy optimization~\cite{hac, rdogaussians}, demonstrating considerable improvements in reducing storage consumption.

However, the aforementioned compression methods typically rely on a large number of Gaussians (over one million). This is due to two major challenges when the number of Gaussians is drastically reduced: 1) each Gaussian needs to represent a larger portion of the scene, making it more susceptible to compression loss, and 2) the increased spacing between Gaussians disrupts spatial locality, leading to higher attribute irregularity and posing challenges for entropy minimization and efficient compression.
Since the number of Gaussians directly impacts computational costs, including training time and rendering speed, it is crucial to develop approaches that effectively minimize the number of Gaussians while maintaining compressibility.

In this paper, we propose Optimized Minimal Gaussian representation (OMG), an efficient compression framework that operates with a minimal number of primitives.
To address the irregularity of sparse Gaussians and maximize the compressibility, we employ per-Gaussian features in a novel way.
Although the reduced number of Gaussians leads to a decrease in local continuity, we can still leverage the spatial correlation associated with each Gaussian’s position.
Therefore, we introduce a lightweight neural field model with negligible parameters to capture the coarse spatial feature.
This feature is integrated with the per-Gaussian features to represent each attribute, as shown in \Cref{fig:arch}.
This approach requires fewer per-Gaussian parameters than directly learning the original attributes, enabling a more compact representation.
While the proposed OMG architecture effectively represents sparse Gaussians, the use of per-Gaussian features impacts storage efficiency. To mitigate this, we introduce a sub-vector quantization (SVQ, \Cref{fig:svq}(c)), which splits the input vector into multiple sub-vectors and applies vector quantization to each sub-vector. This approach alleviates the computational overhead associated with large vector quantization codebooks (\Cref{fig:svq}(a)) and reduces the storage burden caused by the multiple indexing stages of residual vector quantization (\Cref{fig:svq}(b)), while maintaining high-precision representation.

Finally, to retain only the minimal number of Gaussians, we introduce a novel importance metric that evaluates each Gaussian’s local distinctiveness relative to its neighbors, identifying the most informative Gaussians. This metric is used alongside existing importance scoring methods based on blending weights from training views~\cite{radsplat, msv1}, further reducing the number of Gaussians while preserving scene fidelity.

Extensive experimental results demonstrate that OMG achieves a 49\% reduction in storage compared to the previous state-of-the-art method~\cite{locogs}, requiring only 4.1MB for the Mip-NeRF 360 dataset~\cite{mip360} while preserving comparable rendering quality. Additionally, OMG utilizes only 0.4 million Gaussians, enabling 600+ FPS rendering. These results underscore the effectiveness of OMG in both compression efficiency and computational performance, demonstrating it as a highly promising approach for 3D Gaussian Splatting representation.

\section{Related work}
\subsection{Neural Radiance Fields}
Neural Radiance Fields (NeRF)~\cite{nerf} introduced a pioneering approach for novel view synthesis by leveraging volumetric rendering in conjunction with multi-layer perceptrons (MLPs) to model continuous 3D scenes. While NeRF achieves high-quality rendering, its reliance on MLP leads to inefficiencies, particularly in terms of slow training and inference times. To overcome these limitations, the following methods~\cite{fridovich2022plenoxels, nsvf} utilized explicit voxel-based representations, enabling significantly faster training compared to traditional MLP-based NeRF models. However, these approaches still suffer from slow inference speeds and impose substantial memory requirements, posing challenges for scalability and practical deployment in large-scale environments.

\noindent\textbf{Compact representation.} 
To mitigate the memory overhead while maintaining rendering fidelity, various works have been introduced, including grid factorization~\cite{tensorf, EG3D, kplanes, multensorf, strivec}, hash grids~\cite{instant-ngp, howfar}, grid quantization~\cite{VQAD, birf}, and pruning-based strategies~\cite{masked}. Nevertheless, achieving real-time rendering for complex, large-scale scenes remains a formidable challenge. The fundamental limitation of these approaches stems from the necessity of dense volumetric sampling, which, despite optimizations, continues to constrain training and inference speed.

\subsection{3D Gaussian Splatting}
Recently, 3D Gaussian Splatting (3DGS)~\cite{3dgs} has emerged as a paradigm-shifting technique for real-time neural rendering by representing a scene with 3D Gaussian primitives. 3DGS leverages customized CUDA kernels and optimized algorithms to achieve unparalleled rendering speed while preserving high image quality. Unlike volumetric methods that require dense per-ray sampling, 3DGS projects Gaussians onto the image plane and rasterizes them tile-wise, significantly improving computational efficiency. Due to its versatility, 3DGS has become a dominant paradigm in 3D representation, leading to advancements across various domains and applications, such as mesh extraction~\cite{sugar, 2dgs}, simultaneous localization and mapping (SLAM)~\cite{splatam}, dynamic scene representation~\cite{dynamic3dgs}, multi-resolution rendering~\cite{mipsplatting}, and further improvements in rendering quality~\cite{msv1}. However, 3DGS requires a substantial number of Gaussians to maintain high-quality rendering. Furthermore, each primitive is represented with multiple attributes, such as covariance matrices and spherical harmonics (SH) coefficients, requiring a large number of learnable parameters. Consequently, 3DGS demands substantial memory and storage resources, often exceeding 1GB per scene in high-resolution environments.

\noindent\textbf{Reducing the number of primitives.} 
To alleviate the substantial computational and memory overhead of 3DGS, numerous methods have been proposed to reduce the number of Gaussians while preserving rendering quality. Several approaches follow 3DGS by pruning low-opacity Gaussians, incorporated with opacity regularization~\cite{compact3d}, anchored Gaussians~\cite{scaffoldgs}, or hyperparameter search~\cite{c3d2d}.
An alternative approach utilized binary masking techniques~\cite{c3dgs, hac, f-3dgs, rdogaussians, contextgs}, where pruning decisions are directly learned based on rendering loss. To optimize the binary masks, Compact-3DGS~\cite{compact3d} initially adopted STE~\cite{ste}, while subsequent works~\cite{lp-3dgs, maskgaussian} employed Gumbel-Softmax.

Another direction focuses on importance-based metrics to identify and remove redundant Gaussians. These methods primarily leverage each Gaussian’s blending weight contribution to rendering training-view images as a measure of importance~\cite{lightgaussian, msv1, radsplat, eagles, taming3dgs}. LightGaussian~\cite{lightgaussian} further incorporates Gaussian volume and opacity into the importance computation, while Taming 3DGS~\cite{taming3dgs} integrates multiple information, including gradients, pixel saliency, and Gaussian attributes.
Building upon these advancements, we introduce a novel importance metric that incorporates color distinction among neighboring Gaussians, enabling more effective selection of essential primitives.

\noindent\textbf{Attribute compression.}
Earlier methods employed conventional compression techniques such as scalar and vector quantization (VQ)~\cite{lightgaussian, compact3d, morton3d, c3d2d, eagles, reduced3dgs, mesongs} and entropy coding~\cite{morton3d, c3d2d, c3dgs, hac, ff3d} to reduce storage requirements. VQ-based representations have proven highly efficient by the fact that many Gaussian attributes are redundant across a scene, allowing for compact encoding. However, a large codebook leads to substantial computational overhead, increasing training time. While residual vector quantization (R-VQ)~\cite{c3dgs} can alleviate computational costs, it introduces additional storage inefficiencies due to the need for multiple code indices.

Another line of work explored structured representations, incorporating anchor-based encoding~\cite{scaffoldgs, hac, contextgs, compgs} and factorization techniques~\cite{f-3dgs}, integrated with grid representations. Scaffold-GS~\cite{scaffoldgs} first introduced an anchor-based approach, where attributes of grouped Gaussians are encoded using shared anchor features and MLP-based refinements. Building upon this, subsequent methods~\cite{hac, hac++, contextgs, compgs} incorporated context modeling to further improve compression rates. While showing high compression performance, these approaches require per-view processing, involving multiple MLP forward passes, which results in significant rendering latency.

Recent efforts have utilized neural field architectures to exploit the local continuity of neighboring Gaussians. Compact-3DGS~\cite{c3dgs} encodes view-dependent color, while LocoGS~\cite{locogs} represents all Gaussian attributes except for view-independent color. However, unlike NeRF-based representations, where exact spatial positions are used as inputs to neural fields, mapping Gaussian center points to their corresponding attributes remains challenging. This difficulty leads to the use of large neural field models to achieve accurate reconstruction.
In this work, we propose a novel approach that effectively captures both the continuity and irregularities across Gaussians, enabling a more efficient and compact attribute representation.

\section{Method}
\noindent\textbf{Background.} 3DGS represents a scene using a set of $N$ Gaussians, parameterized by their attributes: center position $p \in \mathbb{R}^{N\times3}$, opacity $o \in [0,1]^{N}$, 3D scale $s \in \mathbb{R}_{+}^{N\times3}$, 3D rotation represented as a quaternion $r \in\mathbb{R}^{N\times4}$, and view-dependent color modeled using spherical harmonics (SH) coefficients $h^{(0)}\in\mathbb{R}^{N\times3}$ (0 degree for static color), $h^{(1,2,3)}\in\mathbb{R}^{N\times45}$ (1 to 3 degrees for view-dependent color). The covariance matrix of each Gaussian $\Sigma_n \in \mathbb{R}^{3\times3}$ is determined by scale $s_n$ and rotation $r_n$ attributes.

To render an image, 3D Gaussians are projected into 2D space. Each pixel color in the image $C(\cdot)$ is then rendered through the alpha composition using colors $c_n$ (determined by spherical harmonics under the given view direction) and the final opacity in 2D space $\alpha_n(\cdot)$,
\begin{align}
\label{eq:rast}
    &C(x) = \sum_{k=1}^{\mathcal{N}(x)}c_k\alpha_k(x)\prod_{j=1}^{k-1}(1-\alpha_j(x)),\\
    &\alpha_n(x) = {o_n}\operatorname*{exp}\left(-{1\over2}(x-p'_n)^T{\Sigma'}_n^{-1}(x-p'_n)\right),
\label{eq:projopa}
\end{align}
where $x$ denotes a pixel coordinate and $\Sigma'_n, p'_n$ are the projected Gaussian covariance and center position. $\mathcal{N}(x)$ represents the number of Gaussians around $x$, where the Gaussians are depth-sorted based on the given viewing direction.



\subsection{Overall architecture}
OMG is designed to accurately and efficiently represent the attributes of the minimal Gaussian primitives. Existing approaches~\cite{c3dgs, locogs} have leveraged neural fields to exploit the local continuity of Gaussian attributes. While effective in dense representations, this assumption weakens as Gaussians become sparser. In sparse settings, neighboring Gaussians are further apart, and smooth transitions between them become insufficient to maintain fidelity. Especially for geometry, each Gaussian covers a larger spatial region, requiring a more specific scale and rotation to accurately capture structural details. Therefore, we retain the per-Gaussian parameterization for scale $s \in \mathbb{R}_+^{N \times 3}$ and rotation $r \in \mathbb{R}^{N \times 4}$ as in 3DGS. 

\begin{figure*}[t]
    \begin{center}
    \includegraphics[width=1.0\linewidth]{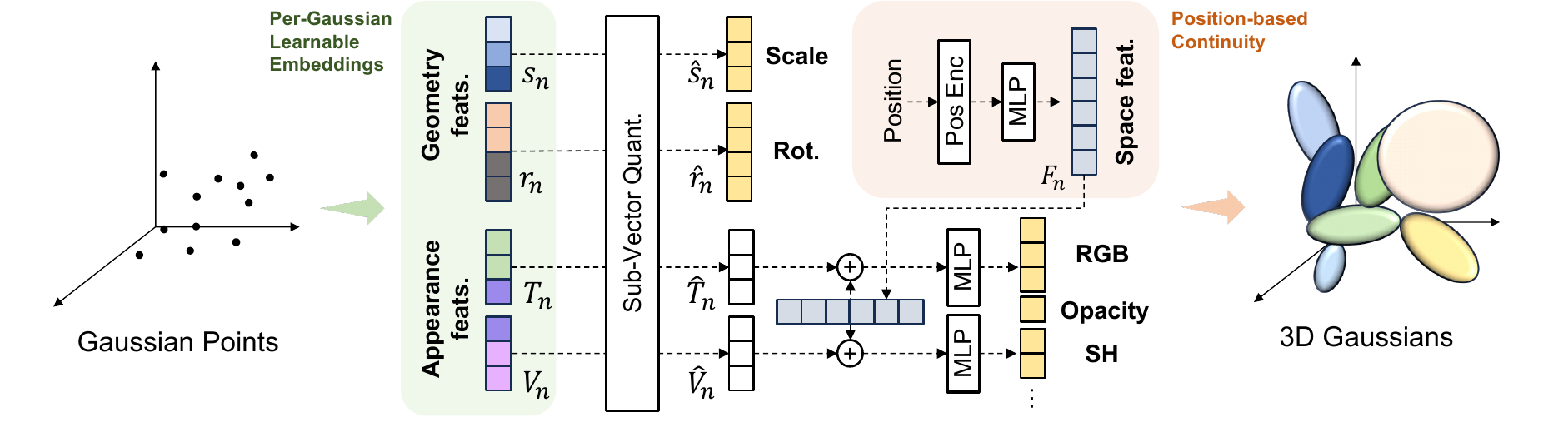}
    \end{center}
    \vspace{-1em}
    \caption{The overall architecture of our proposed OMG. OMG learns per-Gaussian geometric and appearance features, applying Sub-Vector Quantization (SVQ) to all of them. The SVQ-applied geometric attributes are used for rendering, while the space feature based on the Gaussian center position is integrated into the appearance features to define the final appearance.}
    \vspace{-0.5em}
\label{fig:arch}
\end{figure*}

For appearance, local continuity can still be maintained even with increased sparsity.
However, unlike NeRF, where the query input is a direct spatial point, mapping Gaussian center points to corresponding appearances is inherently challenging. This requires a larger neural field model to maintain high fidelity. On the other hand, entirely disregarding local continuity leads to an inefficient representation, limiting the ability to capture meaningful spatial relationships. 
OMG addresses these challenges by integrating per-Gaussian attributes with a small neural field structure, effectively leveraging both irregularity and continuity.

To represent appearance, we learn per-Gaussian attributes, namely a static feature $T \in \mathbb{R}^{N \times D}$ and a view-dependent feature $V \in \mathbb{R}^{N \times D}$, where $D$ is the dimensionality of each feature.
As illustrated in \Cref{fig:arch}, each feature is concatenated with the space feature $F_n$, derived from each Gaussian's center position, to generate static and view-dependent color, and opacity. The space feature itself is efficiently parameterized using positional encoding and an MLP, ensuring a highly compact representation.  
Formally, this process can be expressed as follows:
\begin{align}
\label{eq:static}
&h_n^{(0)} = \text{MLP$_t$}\big(\text{cat}(T_n, F_n)),\, o_n = \text{MLP$_o$}\big( \text{cat} (T_n, F_n)),\\
&h_n^{(1,2,3)} = \text{MLP$_v$}\big(\text{cat}(V_n, F_n)\big),\, F_n = \text{MLP$_s$}\big( \gamma(p_n)\big),
\label{eq:view}
\end{align}
where $\text{cat}(\cdot,\cdot)$ denotes the concatenation function, $\gamma(\cdot)$ represents the positional encoding function, and $\text{MLP$_t$}(\cdot), \text{MLP$_o$}(\cdot), \text{MLP$_v$}(\cdot), \text{MLP$_s$}(\cdot)$ are the MLPs for static color, opacity, view-dependent color, and space feature, respectively.

\subsection{Sub-vector quantization}
\begin{figure}[t]
    \begin{center}
    \includegraphics[width=1.0\linewidth]{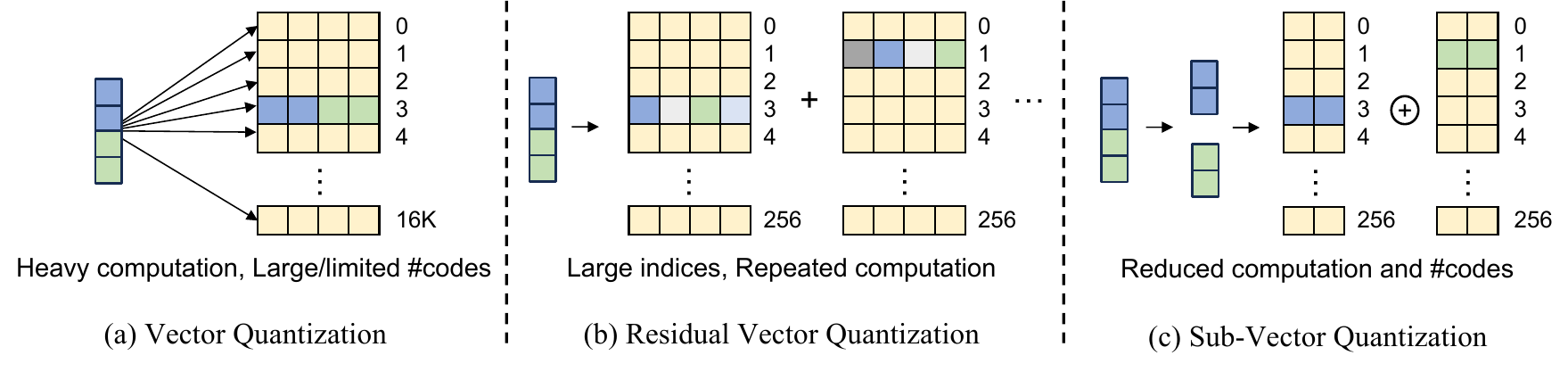}
    \end{center}
    \vspace{-1em}
    \caption{Conceptual diagram of (a) vector quantization, (b) residual vector quantization, and (c) sub-vector quantization. + and $\oplus$ denote the element-wise summation and the vector concatenation.}
    \vspace{-1em}
\label{fig:svq}
\end{figure}

Vector quantization (VQ)~\cite{vq} has shown high efficiency for representing Gaussian attributes, capitalizing on their inherently vectorized structure and strong global coherence across an entire scene. However, to maintain high fidelity, a large codebook size is required, inevitably resulting in substantial computational overhead and increased training complexity~\cite{soundstream} (\Cref{fig:svq}(a)).
To address these issues, Residual Vector Quantization (R-VQ)~\cite{soundstream} has been used as a hierarchical quantization strategy~\cite{c3dgs}, progressively refining representations while reducing the size of each individual codebook.
However, as shown in \Cref{fig:svq}(b), multiple code indices per attribute result in increased storage overhead, illustrating a tradeoff between reducing per-codebook complexity and increasing overall storage requirements.

To navigate this tradeoff, we propose Sub-Vector Quantization (SVQ), which partitions the attribute vector into multiple sub-vectors and applies vector quantization separately to each component (\Cref{fig:svq}(c)), motivated by Product Quantization~\cite{pq}.
By reducing the dimensionality of each quantized unit, SVQ allows for smaller codebooks and more efficient lookups, which can balance computational cost and storage efficiency while maintaining high fidelity.
We can apply SVQ to an input vector $z \in \mathbb{R}^{M L}$, where $M$ and $L$ represent the total number of sub-vectors (partitions) and the sub-vector length, respectively.
Each partition $m \in \{1, ..., M\} $ has an independent codebook $ C^{(m)} \in \mathbb{R}^{B \times L}$, where $B$ denotes the number of codewords per codebook. The codeword selection is based on the nearest match from $C^{(m)}$, with $C^{(m)}[j]$ representing the $j$-th codeword corresponding to the $m$-th sub-vector. More formally, SVQ-applied vector $\hat{z}$ can be formulated as follows,
\begin{align}
&\hat{z} := q(z; M) = \text{cat}(C^{(1)}[i_1], C^{(2)}[i_2], ..., C^{(M)}[i_M]),\\
&i_m = \arg\min_{j} \| z_m - C^{(m)}[j] \|^2_2, \quad m \in \{1, ..., M\},
\end{align}
where $q(z; M)$ denotes applying SVQ with M sub-vectors and $i_m \in \{1,…,B\}$ is the selected index of $m$-th sub-vector. 

SVQ ensures significantly reduced computation with small codebooks compared to VQ. We apply SVQ to geometric attributes $s_n, r_n$, resulting in quantized vectors $\hat{s}_n, \hat{r}_n$, which are then used for 3DGS rendering. For the appearance features $T_n, V_n$, we first concatenate them and apply SVQ. The resulting quantized features are then split back into two components $\hat{T}_n, \hat{V}_n$, which replace $T_n$ and $V_n$ in \Cref{eq:static,eq:view}.

Although the reduced codebook size significantly decreases computational overhead compared to VQ, the process of updating both the indices and codes at every training iteration increases training time. Moreover, we observe that as training converges, the selected codebook indices remain largely unchanged.
Therefore, we adopt a fine-tuning strategy in the final 1K iterations: after initializing with K-means, we freeze the indices and finetune only the codebook using the rendering loss, without introducing any additional losses. Since K-means initialization is completed within seconds due to the small codebooks, this approach adds minimal additional training time, unlike other methods that incur significant overhead.

\subsection{Local distinctiveness for important scoring}
OMG adopts importance scoring to identify essential Gaussians and retain a minimal number of them.
Existing scoring-based pruning methods typically determine the importance of each Gaussian based on its blending weights (the values multiplied by $c_k$ in \Cref{eq:rast}) across training-view renderings. We use two factors as our baseline metric: (1) whether it has been the most dominant contributor for at least one ray~\cite{msv1,msv2} and (2) its total blending weight contribution across all training rays~\cite{radsplat, eagles}. Formally, we define the base importance score $\bar{I}$ as:
\begin{align}
    \bar{I}_i =
\begin{cases}
    \sum_{\rho=1}^{N^R} w_{i,\rho}, & \text{if } \exists \rho \in \{1, ..., N^R\} | w_{i,\rho} = \max_{j} w_{j,\rho}, \\
    0, & \text{otherwise},
\end{cases}
\end{align}
where $w_{i,\rho}$ represents the blending weight of Gaussian $i$ for ray $\rho$ and $N^R$ is the number of total rays in training views.

While this score captures global importance, it does not account for redundancy among closely-positioned Gaussians. In cases where multiple Gaussians are located in close proximity, their blending weights tend to be highly similar, thus naively thresholding them can lead to two potential issues: (1) abrupt performance degradation when all similar Gaussians are simultaneously removed, and (2) redundancy when multiple Gaussians with near-identical contributions are retained.

To mitigate these issues, we propose incorporating a local distinctiveness metric into the importance computation. Specifically, we introduce an additional term that measures the similarity of the static appearance feature $T \in \mathbb{R}^{N \times D}$ between neighboring Gaussians, ensuring that locally distinct Gaussians show high importance. The final importance score is defined as:
\begin{align}
I_i = \bar{I}_i  \left( \frac{1}{D} \sum_{j \in \mathcal{N}_i^K} \| T_i - T_j \|_1 \right)^\lambda,
\end{align}
where $\mathcal{N}_i^K$ denotes the set of $K$-nearest neighbors of Gaussian $i$, and $\lambda$ is a scaling factor that adjusts the sensitivity to appearance variation.
As computing exact $K$-nearest neighbors for every Gaussian is computationally expensive, we approximate neighbor selection by sorting Gaussians in Morton order and selecting Gaussians with adjacent indices as their local neighbors.
We remove low-importance Gaussians using CDF-based thresholding~\cite{compress1mb} with a threshold $\tau$.

\begin{figure*}[t]
    \begin{center}
    \vspace{-1em}
    \includegraphics[width=1.0\linewidth]{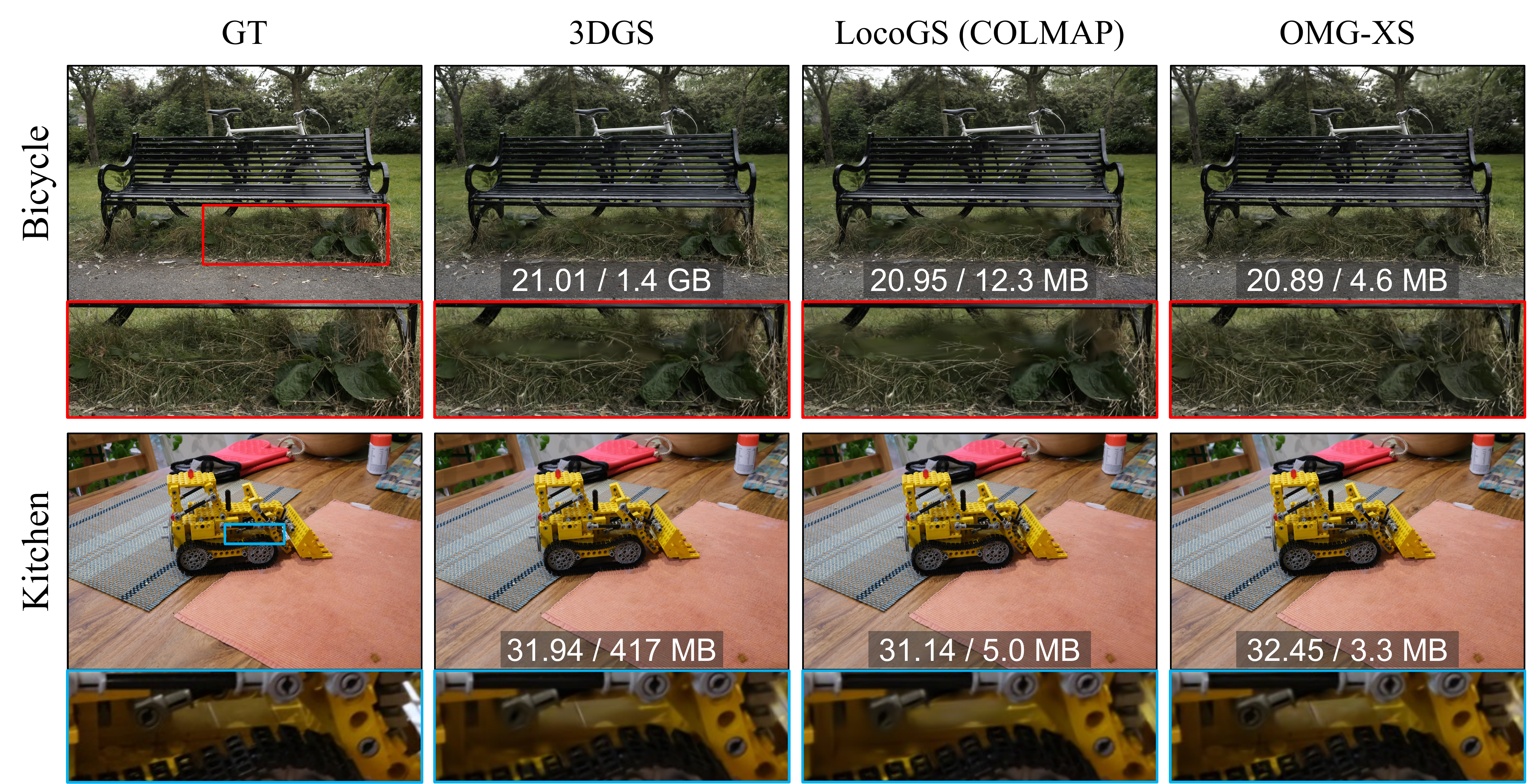}
    \end{center}
    \vspace{-1em}
    \caption{Qualitative results of OMG compared to 3DGS and LocoGS with COLMAP initialization. We provide per-image rendering PSNR with storage requirements for each scene.}
\label{fig:qual}
\end{figure*}

\section{Experiment}

\subsection{Implementation details}
Following the previous works, we evaluated our approach on three real-world datasets, Mip-NeRF 360~\cite{mip360}, Tanks\&Temples~\cite{tnt}, and Deep Blending~\cite{db}.
Our model is implemented upon Mini-Splatting~\cite{msv1}, one of the methods achieving high performance with a small number of Gaussians.
We have conducted simple post-processings after training: 1) Applying 16-bit quantization to the position and compressing with G-PCC~\cite{gpcc}. 2) Huffman encoding~\cite{huffman} to SVQ indices. 3) Storing all the components into a single file with LZMA~\cite{lzma} compression.
We provide five OMG variants (XS, X, M, L, XL), adjusting storage requirements. 
The only factor controlling the storage is the CDF-based threshold value \(\tau\) of Gaussian importance, which is set to 0.96, 0.98, 0.99, 0.999, and 0.9999 for each variant, respectively.
Further implementation details are provided in the supplementary materials.

\begin{table*}[t]
\caption{Quantitative results of OMG evaluated on the Mip-NeRF 360 dataset. Baseline results are sourced from the LocoGS~\cite{locogs} paper, where the rendering results were obtained using an NVIDIA RTX 3090 GPU. Our rendering performance was measured using the same GPU, with the values in parentheses obtained from an NVIDIA RTX 4090 GPU. LocoGS* refers to LocoGS initialized with COLMAP, instead of dense Nerfacto initialization. We highlight the results among compression methods by coloring the \colorbox[HTML]{FFCCC9}{best}, \colorbox[HTML]{FFCE93}{second-best}, and \colorbox[HTML]{FFFC9E}{third-best} performances.}
\label{tab:mip}
\centering
\resizebox{0.74\textwidth}{!}{
\begin{tabular}{lccccc}\toprule
 & \multicolumn{5}{c}{Mip-NeRF 360} \\\cmidrule(lr){2-6}
\multirow{-2}{*}{Method} & PSNR $\uparrow$ & SSIM $\uparrow$ & LPIPS $\downarrow$ & Size(MB) $\downarrow$ & FPS $\uparrow$ \\\midrule
3DGS & 27.44 & 0.813 & 0.218 & 822.6 & 127 \\
Scaffold-GS~\cite{scaffoldgs} & 27.66 & 0.812 & 0.223 & 187.3 & 122 \\
Mini-Splatting~\cite{msv1} & 27.39 & 0.822 & 0.216 & 119.5 & (601) \\\cmidrule(lr){1-1}\cmidrule(lr){2-4}\cmidrule(lr){5-6}
CompGS~\cite{compact3d} & 27.04 & 0.804 & 0.243 & 22.93 & 236 \\
Compact-3DGS~\cite{c3dgs} & 26.95 & 0.797 & 0.244 & 26.31 & 143 \\
C3DGS~\cite{morton3d} & 27.09 & 0.802 & 0.237 & 29.98 & 134 \\
LightGaussian~\cite{lightgaussian} & 26.90 & 0.800 & 0.240 & 53.96 & 244 \\
EAGLES~\cite{eagles} & 27.10 & 0.807 & 0.234 & 59.49 & 155 \\
SOG~\cite{c3d2d} & 27.01 & 0.800 & 0.226 & 43.77 & 134 \\
HAC~\cite{hac} & \cellcolor[HTML]{FFCCC9}27.49 & 0.807 & 0.236 & 16.95 & 110 \\
LocoGS-S~\cite{locogs} & 27.04 & 0.806 & 0.232 & 7.90 & \cellcolor[HTML]{FFCE93}310 \\
LocoGS-L~\cite{locogs} & 27.33 & \cellcolor[HTML]{FFCE93}0.814 & \cellcolor[HTML]{FFCC67}0.219 & 13.89 & 270 \\
\multirow{2}{*}{\begin{tabular}[c]{@{}l@{}}LocoGS* (COLMAP)\end{tabular}} & 27.09 & 0.798 & 0.250 & 7.96 & (396) \\
 & \cellcolor[HTML]{FFCE93}27.37 & 0.807 & 0.236 & 15.10 & (325) \\\cmidrule(lr){1-1}\cmidrule(lr){2-4}\cmidrule(lr){5-6}
OMG-XS & 27.06 & 0.807 & 0.243 & \cellcolor[HTML]{FFCCC9}4.06 & \cellcolor[HTML]{FFCCC9}350 (612) \\
OMG-M & 27.21 & \cellcolor[HTML]{FFCE93}0.814 & \cellcolor[HTML]{FFFC9E}0.229 & \cellcolor[HTML]{FFCE93}5.31 & \cellcolor[HTML]{FFFC9E}298 (511) \\
OMG-XL & \cellcolor[HTML]{FFFC9E}27.34 & \cellcolor[HTML]{FFCCC9}0.819 & \cellcolor[HTML]{FFCCC9}0.218 & \cellcolor[HTML]{FFFC9E}6.82 & 251 (416) \\\bottomrule
\end{tabular}}
\vspace{-1em}
\end{table*}

\begin{table}[t]
\caption{Quantitative results of OMG evaluated on the Tanks\&Temples and Deep Blending datasets. Baseline results are sourced from the LocoGS~\cite{locogs} paper, where the rendering results were obtained using an NVIDIA RTX 3090 GPU. Our rendering performance was measured using the same GPU, with the values in parentheses obtained from an NVIDIA RTX 4090 GPU.}
\label{tab:tntdb}
\centering
\resizebox{\linewidth}{!}{
\begin{tabular}{lcccccccccc}\toprule
 & \multicolumn{5}{c}{Tank\&Temples} & \multicolumn{5}{c}{Deep Blending} \\\cmidrule(lr){2-6}\cmidrule(lr){7-11}
\multirow{-2}{*}{Method} & PSNR $\uparrow$ & SSIM $\uparrow$ & LPIPS $\downarrow$ & Size $\downarrow$ & FPS $\uparrow$ & PSNR $\uparrow$ & SSIM $\uparrow$ & LPIPS $\downarrow$ & Size $\downarrow$ & FPS $\uparrow$ \\\midrule
3DGS~\cite{3dgs} & 23.67 & 0.844 & 0.179 & 452.4 & 175 & 29.48 & 0.900 & 0.246 & 692.5 & 134 \\
Scaffold-GS ~\cite{scaffoldgs} & 24.11 & 0.855 & 0.165 & 154.3 & 109 & 30.28 & 0.907 & 0.243 & 121.2 & 194 \\
Mini-Splatting~\cite{msv1} & 23.41 & 0.846 & 0.180 & 67.6 & (1095) & 30.04 & 0.910 & 0.241 & 124.9 & (902) \\\cmidrule(lr){1-1}\cmidrule(lr){2-6}\cmidrule(lr){7-11}
CompGS~\cite{compact3d} & 23.29 & 0.835 & 0.201 & 14.23 & 329 & 29.89 & \cellcolor[HTML]{FFFC9E}0.907 & 0.253 & 15.15 & 301 \\
Compact-3DGS~\cite{c3dgs} & 23.33 & 0.831 & 0.202 & 18.97 & 199 & 29.71 & 0.901 & 0.257 & 21.75 & 184 \\
C3DGS~\cite{morton3d} & 23.52 & 0.837 & 0.188 & 18.58 & 166 & 29.53 & 0.899 & 0.254 & 24.96 & 143 \\
LightGaussian~\cite{lightgaussian} & 23.32 & 0.829 & 0.204 & 29.94 & \cellcolor[HTML]{FFFC9E}379 & 29.12 & 0.895 & 0.262 & 45.25 & 287 \\
EAGLES~\cite{eagles} & 23.14 & 0.833 & 0.203 & 30.18 & 244 & 29.72 & 0.906 & \cellcolor[HTML]{FFFC9E}0.249 & 54.45 & 137 \\
SOG~\cite{c3d2d} & 23.54 & 0.833 & 0.188 & 24.42 & 222 & 29.21 & 0.891 & 0.271 & 19.32 & 224 \\
HAC~\cite{hac} & \cellcolor[HTML]{FFCCC9}24.08 & \cellcolor[HTML]{FFFC9E}0.846 & 0.186 & 8.42 & 129 & \cellcolor[HTML]{FFFC9E}29.99 & 0.902 & 0.268 & \cellcolor[HTML]{FFCE93}4.51 & 235 \\
LocoGS-S~\cite{locogs} & \cellcolor[HTML]{FFFC9E}23.63 & \cellcolor[HTML]{FFCE93}0.847 & \cellcolor[HTML]{FFCE93}0.169 & \cellcolor[HTML]{FFFC9E}6.59 & 333 & \cellcolor[HTML]{FFCE93}30.06 & 0.904 & \cellcolor[HTML]{FFFC9E}0.249 & 7.64 & \cellcolor[HTML]{FFFC9E}334 \\
LocoGS-L~\cite{locogs} & \cellcolor[HTML]{FFCE93}23.84 & \cellcolor[HTML]{FFCCC9}0.852 & \cellcolor[HTML]{FFCCC9}0.161 & 12.34 & 311 & \cellcolor[HTML]{FFCCC9}30.11 & 0.906 & \cellcolor[HTML]{FFCCC9}0.243 & 13.38 & 297 \\\cmidrule(lr){1-1}\cmidrule(lr){2-6}\cmidrule(lr){7-11}
OMG-M~ & 23.52 & 0.842 & 0.189 & \cellcolor[HTML]{FFCCC9}3.22 & \cellcolor[HTML]{FFCCC9}555 (887) & 29.77 & \cellcolor[HTML]{FFCE93}0.908 & 0.253 & \cellcolor[HTML]{FFCCC9}4.34 & \cellcolor[HTML]{FFCCC9}524 (894) \\
OMG-L & 23.60 & \cellcolor[HTML]{FFFC9E}0.846 & \cellcolor[HTML]{FFFC9E}0.181 & \cellcolor[HTML]{FFCE93}3.93 & \cellcolor[HTML]{FFCE93}478 (770) & 29.88 & \cellcolor[HTML]{FFCCC9}0.910 & \cellcolor[HTML]{FFCE93}0.247 & \cellcolor[HTML]{FFFC9E}5.21 & \cellcolor[HTML]{FFCE93}479 (810) \\\bottomrule
\end{tabular}}
\end{table}

\subsection{Performance evaluation}
\noindent\textbf{Compression performance. }\Cref{tab:mip,tab:tntdb} compare the performance of OMG against various baseline methods on the Mip-NeRF 360, Tanks \& Temples, and DeepBlending datasets. OMG consistently shows the smallest storage requirements while maintaining high performance across all datasets, achieving state-of-the-art (SOTA) results.
Notably, on the Mip-NeRF 360 dataset, OMG-XS achieves nearly a 50\% reduction in storage compared to the small variant of LocoGS~\cite{locogs}, the previous SOTA compression method, while retaining PSNR and SSIM. With over 30\% reduced storage, OMG-M outperforms LocoGS-S in all quality metrics. Moreover, OMG-XL surpasses LocoGS-L in all metrics, even though requiring less storage than LocoGS-S.

The qualitative results presented in \Cref{fig:qual} also demonstrate the strong performance of OMG. Despite achieving over 100× compression compared to 3DGS, OMG maintains comparable visual quality.
Especially, in the \textit{bicycle} scene, OMG-XS achieves over 300× compression relative to 3DGS while accurately reconstructing details that 3DGS fails to represent, resulting in a blurry area (highlighted in red) in its rendering.  This superiority can be attributed to the blur split technique of our baseline model, Mini-Splatting~\cite{msv1}. Despite reducing the number of Gaussians by an additional 20\% compared to Mini-Splatting (\Cref{tab:eff}), OMG-XS retains high visual fidelity, demonstrating its effectiveness in extreme compression scenarios.

\noindent\textbf{Computational efficiency. }
OMG achieves remarkable efficiency alongside high performance. As shown in \Cref{tab:eff}, OMG shows superior scene fidelity with significantly fewer Gaussian primitives compared to LocoGS. This reduction results in substantial rendering speed improvements of 13\%, 67\%, and 57\% for the Mip-NeRF 360, Tank\&Temples, and Deep Blending datasets (\Cref{tab:mip,tab:tntdb}), respectively, compared to LocoGS, highlighting its potential for real-time rendering on low-capacity devices.
Furthermore, OMG accelerates training speed. The substantial improvement over LocoGS can be attributed to two key factors: the reduced number of Gaussians and the absence of a large neural field. By efficiently exploiting coarse spatial information through a tiny MLP, OMG achieves high computational efficiency.

\begin{table}[t]
\centering
\caption{Efficiency comparison of OMG variants evaluated on the Mip-NeRF 360 dataset. We present training time, the number of Gaussians, and the storage requirement with rendering quality.}
\label{tab:eff}
\resizebox{0.66\textwidth}{!}{
\begin{tabular}{lcccccc}\toprule
\multicolumn{1}{l}{Method} & Training & \#Gauss & Size & PSNR & SSIM & LPIPS \\\midrule
Mini-Splatting & 19m 25s & 531K & 119.5 & 27.39 & 0.822 & 0.216 \\
\cmidrule(lr){1-1}\cmidrule(lr){2-4}\cmidrule(lr){5-7}
LocoGS-S & \multirow{2}{*}{\begin{tabular}[c]{@{}c@{}}$\approx$1h\end{tabular}} & 1.09M & 7.9 & 27.04 & 0.806 & 0.232 \\
LocoGS-L &  & 1.32M & 13.89 & 27.33 & 0.814 & 0.219 \\
\cmidrule(lr){1-1}\cmidrule(lr){2-4}\cmidrule(lr){5-7}
OMG-XS & 20m 15s & 427K & 4.06 & 27.06 & 0.807 & 0.243 \\
OMG-S & 20m 57s & 501K & 4.75 & 27.14 & 0.811 & 0.235 \\
OMG-M & 21m 10s & 563K & 5.31 & 27.21 & 0.814 & 0.229 \\
OMG-L & 21m 32s & 696K & 6.52 & 27.28 & 0.818 & 0.220 \\
OMG-XL & 22m 26s & 727K & 6.82 & 27.34 & 0.819 & 0.218 \\\bottomrule
\end{tabular}}
\end{table}


\begin{figure}[t]
\begin{minipage}[]{0.51\textwidth}
\vspace{-0.8em}
    \includegraphics[width=1.0\linewidth]{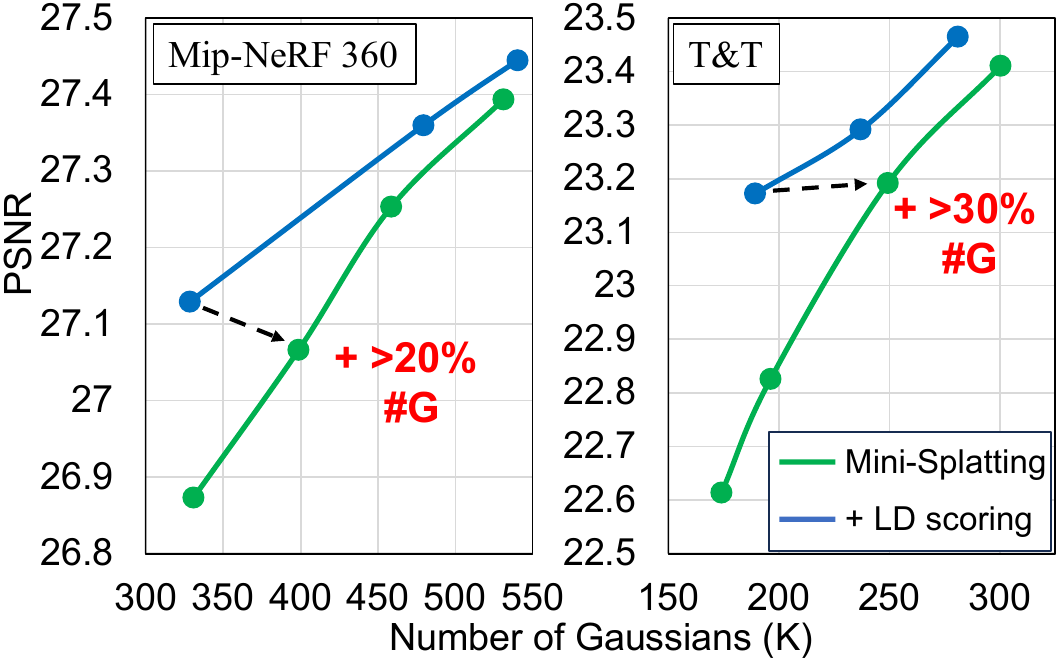}
    \vspace{-1.8em}
    \captionof{figure}{Evaluation without attribute compression.}\vspace{-1em}
\label{fig_numG}
\end{minipage}\hfill
\begin{minipage}[]{0.48\textwidth}\vspace*{-0.8em}
\captionof{table}{Ablation study of OMG using the Mip-NeRF 360 dataset.} 
  \vspace{-0.2em}
  \resizebox{1.0\linewidth}{!}{
    \begin{tabular}{lccccc}\toprule
    Method & PSNR & SSIM & LPIPS & \#Gauss & Size \\\midrule
    OMG-M & 27.21 & 0.814 & 0.229 & 0.56M & 5.31 \\
    w/o Space feature & 26.96 & 0.811 & 0.232 & 0.59M & 5.58 \\
    w/o LD scoring & 27.09 & 0.813 & 0.230 & 0.57M & 5.36 \\
    w/o Both & 26.81 & 0.809 & 0.234 & 0.59M & 5.59 \\\cmidrule(lr){1-1}\cmidrule(lr){2-4}\cmidrule(lr){5-6}
    w/o SVQ & 27.26 & 0.817 & 0.226 & 0.56M & 26.1 \\\midrule
    OMG-XS & 27.06 & 0.807 & 0.243 & 0.43M & 4.06 \\
    w/o Space feature & 26.85 & 0.804 & 0.246 & 0.44M & 4.17 \\
    w/o LD scoring & 26.83 & 0.804 & 0.246 & 0.43M & 4.12 \\
    w/o Both & 26.52 & 0.798 & 0.252 & 0.45M & 4.24 \\\cmidrule(lr){1-1}\cmidrule(lr){2-4}\cmidrule(lr){5-6}
    w/o SVQ & 27.06 & 0.809 & 0.241 & 0.43M & 19.8 \\\bottomrule
    \end{tabular}}\vspace{-1em}
\label{tab:abl}
\end{minipage}
\end{figure}

\subsection{Ablation study}

\noindent\textbf{Local distinctiveness (LD) scoring. }
OMG improves Gaussian pruning by incorporating LD scoring into the importance estimation. Without attribute compression, LD scoring enables high rendering quality with extremely reduced Gaussians, achieving similar performance compared to Mini-Splatting that use 20--30\% more Gaussians, as shown in \Cref{fig_numG}. When applying attribute compression (\Cref{tab:eff}), OMG with LD scoring still outperforms, leading to a significant performance improvement with a similar number of Gaussians. This effect becomes even more pronounced when the target Gaussian number is lower, demonstrating that LD scoring provides an effective approach for further reducing a sparse set of Gaussians.

\noindent\textbf{OMG architecture. }
OMG leverages a highly compact neural field to capture coarse spatial information while reducing the number of learnable parameters per Gaussian. \Cref{tab:abl} validates the contribution of this space feature. Although the total number of Gaussians slightly increases, performance significantly degrades. The absence of spatial information introduces instability in attribute learning, hindering effective importance scoring. This trend is consistently observed in both our small and medium models, highlighting the effectiveness of the space feature despite its minimal parameter overhead.
Furthermore, when both the space feature and LD scoring are removed, the model experiences the most substantial performance drop. This indicates that the two contributions are orthogonal, independently contributing to model efficiency and performance.

\begin{table}[t]
\centering
\caption{Ablation study of SVQ on the Mip-NeRF 360 dataset. We replace SVQ with either VQ or RVQ. `CB Init' denotes the time (in seconds) required for codebook initialization. }
\label{tab:svq}
\resizebox{\linewidth}{!}{
\begin{tabular}{lccccccccc} \toprule
Method & \multicolumn{3}{c}{OMG-XS} & \multicolumn{3}{c}{OMG-M} & \multicolumn{3}{c}{OMG-XL} \\\cmidrule(lr){1-1}\cmidrule(lr){2-4}\cmidrule(lr){5-7}\cmidrule(lr){8-10}
Metric & CB Init$\downarrow$ & Size$\downarrow$ & PSNR$\uparrow$ & CB Init$\downarrow$ & Size$\downarrow$ & PSNR$\uparrow$ & CB Init$\downarrow$ & Size$\downarrow$ & PSNR$\uparrow$ \\\midrule
OMG & 4.8 & 4.06 & 27.06 & 7.6 & 5.31 & 27.21 & 7.9 & 6.82 & 27.34 \\
SVQ$\rightarrow$VQ & 86.4 & 4.21 & 27.06 & 100.9 & 5.36 & 27.18 & 117.4 & 6.70 & 27.27 \\
SVQ$\rightarrow$RVQ & 4.0 & 4.19 & 27.01 & 5.5 & 5.52 & 27.13 & 6.9 & 7.10 & 27.25 \\ \bottomrule
\end{tabular}}
\vspace{-1em}
\end{table}

\noindent\textbf{Sub-vector quantization. }
\Cref{tab:abl} shows the effectiveness of SVQ in terms of performance-storage efficiency. In addition, we replaced it with VQ and RVQ while keeping the proposed clever training strategy: K-means clustering is performed once prior to the final 1K iterations, after which only the codebook is updated. As shown in \Cref{tab:svq}, VQ incurs a 13–18$\times$ increase in codebook initialization time compared to SVQ, resulting in significantly higher training overhead due to the need for large codebooks ($2^{14}$ entries per scale, rotation, and two appearance features) to ensure accurate representation. Nevertheless, VQ leads to lower rendering quality and/or increased storage costs across variants.
In contrast, RVQ achieves slightly faster codebook initialization than SVQ but performs poorly in terms of rate-distortion efficiency, yielding higher storage requirements and lower rendering quality. These results demonstrate that SVQ outperforms both VQ and RVQ in representing per-Gaussian attributes efficiently.

\noindent\textbf{Generalization ability. }
The OMG attribute representation is applicable to all 3DGS-based methods, as it flexibly represents attributes with spatial locality (via space features and SVQ) as well as those without (using SVQ alone). To validate this generalizability, we applied OMG to 3DGS-MCMC~\cite{mcmc}, a method well-known for effective densification. As reported in \Cref{tab:mcmc}, OMG preserves the original performance and reduce storage requirements, demonstrating its broader applicability.

\begin{table}[t]
\centering
\caption{Results of applying OMG representation to 3DGS-MCMC~\cite{mcmc} on the Mip-NeRF dataset.}
\resizebox{\linewidth}{!}{
\begin{tabular}{lcccccccccc}
\toprule
Method & \#Gauss & PSNR & SSIM & LPIPS & Size & \#Gauss & PSNR & SSIM & LPIPS & Size \\
\cmidrule(lr){1-1}\cmidrule(lr){2-6}\cmidrule(lr){7-11}
MCMC~\cite{mcmc}   & 500K & 27.42 & 0.807 & 0.248 & 115 MB & 1M   & 27.83 & 0.823 & 0.221 & 230 MB \\
MCMC+OMG    & 500K & 27.21 & 0.797 & 0.256 & 5.1 MB & 1M   & 27.63 & 0.813 & 0.227 & 10.0 MB \\
\bottomrule
\end{tabular}}
\label{tab:mcmc}
\end{table}

\begin{table}[t!]
\centering
\vspace{-1em}
\caption{Performance evaluation using Zip-NeRF~\cite{zipnerf} dataset. * indicates results reported in the SMERF~\cite{smerf} paper, where 3DGS hyperparameters were tuned for higher performance. We train and report results for 3DGS and Mini-Splatting using their default settings.}
\label{tab:zip}
\resizebox{0.7\linewidth}{!}{
\begin{tabular}{lccccc}\toprule
Method & PSNR & SSIM & LPIPS & \#Gauss & Size \\\midrule
Zip-NeRF*~\cite{zipnerf} & 27.37 & 0.836 & 0.305 & - & 607 MB \\
SMERF*~\cite{smerf} & 27.28 & 0.829 & 0.350 & - & 4.1 GB \\
3DGS*~\cite{3dgs} & 25.50 & 0.810 & 0.369 & - & 212 MB \\\cmidrule(lr){1-1}\cmidrule(lr){2-4}\cmidrule(lr){5-6}
3DGS~\cite{3dgs} & 25.16 & 0.813 & 0.358 & 934K & 210 MB \\
Mini-Splatting~\cite{msv1} & 24.57 & 0.802 & 0.370 & 337K & 75.7 MB \\\cmidrule(lr){1-1}\cmidrule(lr){2-4}\cmidrule(lr){5-6}
MCMC~\cite{mcmc} & 26.37 & 0.838 & 0.318 & 3M & 675 MB \\
MCMC+OMG & 26.69 & 0.839 & 0.310 & 3M & 29.0 MB \\\bottomrule
\end{tabular}}
\end{table}

\section{Limitations and broader applicability}
\textbf{Baseline dependancy. } OMG builds upon 3DGS and Mini-Splatting with the goal of preserving original rendering quality under compact representations. However, its applicability remains inherently bounded by the capabilities of these baselines. For instance, 3DGS and subsequent approaches exhibit clear limitations in capturing the complexity of larger-scale scenes such as the Zip-NeRF~\cite{zipnerf} dataset. As shown in \Cref{tab:zip}, both 3DGS and Mini-Splatting fail to densify a sufficient number of Gaussians, leading to degraded performance. Consequently, OMG inherits these limitations, restricting its applicability in such scenarios.

\textbf{Broader applicability. } On the other hand, as demonstrated in \Cref{tab:mcmc}, OMG can be applied in a generalizable manner across different baseline methods. Based on this strength, we tune 3DGS-MCMC for large-scale scenes and incorporate the OMG representation, achieving superior performance while drastically reducing storage requirements (\Cref{tab:zip}). More broadly, when integrated with ongoing research across diverse domains, OMG has the potential to provide a highly general and efficient representation.

\section{Conclusion}
In this paper, we proposed Optimized Minimal Gaussians (OMG), a novel framework that significantly reduces the number of Gaussian primitives while maximizing compressibility and maintaining high rendering quality. By effectively identifying and preserving locally distinct Gaussians, OMG minimizes the redundancy of Gaussians with minimal loss of visual fidelity. Furthermore, our compact and precise attribute representation, combined with sub-vector quantization, enables efficient exploitation of both continuity and irregularity, ensuring high efficiency.
Experimental results demonstrate that OMG reduces storage requirements by nearly 50\% compared to the previous state-of-the-art method while allowing over 600 FPS rendering performance.
OMG sets a new benchmark for highly efficient 3D scene representations, paving the way for future advancements in real-time rendering on resource-constrained devices.

\section*{Acknowledgements}
This work was supported in part by the Institute of Information and Communications Technology Planning and Evaluation (IITP) grants (RS-2019-II190421, RS-2020-II201821, RS-2021-II212052, RS-2021-II212068, RS-2024-00457882, RS-2025-02217613, RS-2025-10692981, RS-2025-25442569) funded by the Korean government (MSIT); the Technology Innovation Program (RS-2023-00235718, 23040-15FC) funded by the Ministry of Trade, Industry \& Energy (MOTIE, Korea) (1415187505); Samsung Research Funding \& Incubation Center of Samsung Electronics (SRFC-IT2401-01); a grant of the Korea-US Collaborative Research Fund (KUCRF), funded by the Ministry of Science and ICT and Ministry of Health \& Welfare, Republic of Korea (RS-2024-00468417).

{
    \small
    \bibliographystyle{plainnat}
    \bibliography{bibtex}
}


\newpage
\section*{NeurIPS Paper Checklist}

\begin{enumerate}

\item {\bf Claims}
    \item[] Question: Do the main claims made in the abstract and introduction accurately reflect the paper's contributions and scope?
    \item[] Answer: \answerYes{} 
    \item[] Justification: The abstract and introduction accurately reflect the paper's contribution, Optimized Minimal Gaussians (OMG) representation, which reduces storage requirements and enables faster rendering compared to existing 3DGS methods.

\item {\bf Limitations}
    \item[] Question: Does the paper discuss the limitations of the work performed by the authors?
    \item[] Answer: \answerYes{} 
    \item[] Justification: We discuss the limitations of our method in Section 5.

\item {\bf Theory assumptions and proofs}
    \item[] Question: For each theoretical result, does the paper provide the full set of assumptions and a complete (and correct) proof?
    \item[] Answer: \answerNA{} 
    \item[] Justification: The paper does not present formal theoretical results or proofs.

    \item {\bf Experimental result reproducibility}
    \item[] Question: Does the paper fully disclose all the information needed to reproduce the main experimental results of the paper to the extent that it affects the main claims and/or conclusions of the paper (regardless of whether the code and data are provided or not)?
    \item[] Answer: \answerYes{} 
    \item[] Justification: We provide sufficient details for reproducing the main experimental results, including descriptions of the datasets used, evaluation metrics, and implementation details.

\item {\bf Open access to data and code}
    \item[] Question: Does the paper provide open access to the data and code, with sufficient instructions to faithfully reproduce the main experimental results, as described in supplemental material?
    \item[] Answer: \answerYes{} 
    \item[] Justification: The experiments are conducted using open datasets, and the paper provides detailed implementation descriptions. Additionally, the source code is attached in the supplementary material to ensure reproducibility.

\item {\bf Experimental setting/details}
    \item[] Question: Does the paper specify all the training and test details (e.g., data splits, hyperparameters, how they were chosen, type of optimizer, etc.) necessary to understand the results?
    \item[] Answer: \answerYes{} 
    \item[] Justification: The paper specifies training and testing details, including datasets, hyperparameters, and optimizer settings, necessary to understand the results.

\item {\bf Experiment statistical significance}
    \item[] Question: Does the paper report error bars suitably and correctly defined or other appropriate information about the statistical significance of the experiments?
    \item[] Answer: \answerNo{} 
    \item[] Justification: Error bars are not provided, following the common practice in this field where experimental results typically do not vary significantly across runs.

\item {\bf Experiments compute resources}
    \item[] Question: For each experiment, does the paper provide sufficient information on the computer resources (type of compute workers, memory, time of execution) needed to reproduce the experiments?
    \item[] Answer: \answerYes{} 
    \item[] Justification: The paper provides information on the compute resources used, including GPU models (e.g., NVIDIA RTX 3090 and 4090), and reports training times and rendering speeds achieved.
    
\item {\bf Code of ethics}
    \item[] Question: Does the research conducted in the paper conform, in every respect, with the NeurIPS Code of Ethics \url{https://neurips.cc/public/EthicsGuidelines}?
    \item[] Answer: \answerYes{} 
    \item[] Justification: The research adheres to ethical guidelines, with no indication of unethical practices or violations of the NeurIPS Code of Ethics.

\item {\bf Broader impacts}
    \item[] Question: Does the paper discuss both potential positive societal impacts and negative societal impacts of the work performed?
    \item[] Answer: \answerNA{} 
    \item[] Justification: There is no societal impact of the work performed. This paper is focused on optimizing the efficiency of an existing framework. 
    
\item {\bf Safeguards}
    \item[] Question: Does the paper describe safeguards that have been put in place for responsible release of data or models that have a high risk for misuse (e.g., pretrained language models, image generators, or scraped datasets)?
    \item[] Answer: \answerNA 
    \item[] Justification: The paper does not release models or data with high risk for misuse.

\item {\bf Licenses for existing assets}
    \item[] Question: Are the creators or original owners of assets (e.g., code, data, models), used in the paper, properly credited and are the license and terms of use explicitly mentioned and properly respected?
    \item[] Answer: \answerYes{} 
    \item[] Justification: The paper references existing datasets and methods used, respecting their licenses and terms of use.

\item {\bf New assets}
    \item[] Question: Are new assets introduced in the paper well documented and is the documentation provided alongside the assets?
    \item[] Answer: \answerYes{} 
    \item[] Justification: The new assets introduced, including the OMG codebase, are well documented and provided alongside the paper for reproducibility.

\item {\bf Crowdsourcing and research with human subjects}
    \item[] Question: For crowdsourcing experiments and research with human subjects, does the paper include the full text of instructions given to participants and screenshots, if applicable, as well as details about compensation (if any)? 
    \item[] Answer: \answerNA{} 
    \item[] Justification: The paper does not involve crowdsourcing or research with human subjects.

\item {\bf Institutional review board (IRB) approvals or equivalent for research with human subjects}
    \item[] Question: Does the paper describe potential risks incurred by study participants, whether such risks were disclosed to the subjects, and whether Institutional Review Board (IRB) approvals (or an equivalent approval/review based on the requirements of your country or institution) were obtained?
    \item[] Answer: \answerNA{} 
    \item[] Justification: The research does not involve crowdsourcing nor research with human subjects.

\item {\bf Declaration of LLM usage}
    \item[] Question: Does the paper describe the usage of LLMs if it is an important, original, or non-standard component of the core methods in this research? Note that if the LLM is used only for writing, editing, or formatting purposes and does not impact the core methodology, scientific rigorousness, or originality of the research, declaration is not required.
    \item[] Answer: \answerNA{} 
    \item[] Justification:  The research does not involve the use of large language models as a core component of the methodology.

\end{enumerate}

\newpage
\appendix
\section{Difficulty in compressing small set of Gaussians}
To assess the challenge of compressing a small number of Gaussians, we integrate Mini-Splatting~\cite{msv1} with existing methods~\cite{c3dgs, locogs}. We trained LocoGS~\cite{locogs} with Mini-Splatting-based densification in an end-to-end manner. For Compact-3DGS~\cite{c3dgs}, however, we observe that training it directly with Mini-Splatting-based pruning fails to converge, due to inaccurate attribute representations during training. Rather, we halt Mini-Splatting training early at 20K iterations, where the pruning process is ended, and integrate Compact-3DGS representation in the subsequent 10K iterations. 

As shown in \Cref{tab:c3dgs}, the integration of Mini-Splatting results in a significant degradation in visual quality across both methods, primarily caused by imprecise attribute representations for irregular and sparse Gaussians. In contrast, OMG preserves high visual fidelity with drastically reduced storage by through its novel attribute representation.

\begin{table}[h]
\centering
\caption{Performance evaluation compared to Compact-3DGS~\cite{c3dgs} and LocoGS~\cite{locogs} with the reduced number of Gaussians.}
\label{tab:c3dgs}
\begin{tabular}{lccccc}\toprule
Method & PSNR & SSIM & LPIPS & \#G & Size \\\midrule
Mini-Splatting (MS) & 27.39 & 0.822 & 0.216 & 531K & 120 MB \\\cmidrule(lr){1-1}\cmidrule(lr){2-4}\cmidrule(lr){5-6}
Compact-3DGS & 27.03 & 0.797 & 0.247 & 1.39M & 29.1 MB \\
MS + Compact-3DGS & 26.85 & 0.807 & 0.234 & 525K & 14.8 MB \\\cmidrule(lr){1-1}\cmidrule(lr){2-4}\cmidrule(lr){5-6}
LocoGS-S & 27.04 & 0.806 & 0.232 & 1.09M & 7.9 MB \\
MS + LocoGS-S & 26.52 & 0.789 & 0.257 & 537K & 6.2 MB \\
LocoGS-L & 27.33 & 0.814 & 0.219 & 1.32M & 13.9 MB \\
MS + LocoGS-L & 26.76 & 0.799 & 0.247 & 525K & 11.3 MB \\\bottomrule
\end{tabular}
\end{table}

\section{KNN approximation using Morton order}
\Cref{tab:morton} presents a comparison between Morton-order-based approximation and KNN for LD scoring. The performance is nearly identical, demonstrating that Morton order effectively approximates KNN while achieving a 146$\times$ speedup, even with the smallest model.
While using KNN is reasonable in our current setup (LD scoring once at 20K with fewer than 1M Gaussians), our decision to adopt Morton order is motivated by the need for scalability and generalizability. In more complex or larger-scale scenes, the number of Gaussians can increase significantly during earlier training stages. In such cases, KNN becomes impractical due to its high memory usage and computational complexity of $O(N^2)$. Morton order, by contrast, provides a well-approximated yet efficient alternative that scales better with the number of primitives.

\begin{table}[h]
\centering
\caption{Performance evaluation of KNN approximation on the Mip-NeRF 360 dataset.}
\begin{tabular}{lcccc}
\toprule
Method & PSNR & SSIM & LPIPS & Time (ms) \\
\midrule
OMG-XS          & 27.06 & 0.807 & 0.243 & 4100 \\
Morton $\rightarrow$ KNN & 27.07 & 0.808 & 0.241 & 28 \\
\bottomrule
\end{tabular}
\label{tab:morton}
\end{table}

\section{Rendering speed on weaker devices}
We have measured the FPS of our method on an NVIDIA RTX 4090 and a lower-end GPU, NVIDIA GTX 1080 Ti. As shown in \Cref{tab:fps}, OMG achieves substantially higher FPS on both devices compared to LocoGS, primarily due to its drastically reduced number of Gaussians.
Especially, OMG-XS show 1.5× faster rendering on the GTX 1080Ti while requiring only half the storage and preserving high visual quality.
These results indicate that OMG’s efficiency directly extends to practical real-time applications, enabling deployment on lower-end GPUs.

\begin{table}[t]
\centering
\caption{Rendering FPS evaluated on the RTX 4090 and GTX 1080Ti, compared to LocoGS (with COLMAP initialization).}
\begin{tabular}{lccccccc}
\toprule
Method & 4090 & 1080Ti & PSNR & SSIM & LPIPS & \#Gauss & Size (MB) \\
\midrule
LocoGS & 396 & 73  & 27.09 & 0.798 & 0.250 & 1.04M & 7.96 \\
OMG-XS & 612 & 106 & 27.06 & 0.807 & 0.243 & 0.43M & 4.06 \\
\cmidrule(lr){1-1}\cmidrule(lr){2-3}\cmidrule(lr){4-6}\cmidrule(lr){7-8}
LocoGS & 325 & 59  & 27.37 & 0.807 & 0.236 & 1.44M & 15.1 \\
OMG-XL & 416 & 77  & 27.34 & 0.819 & 0.218 & 0.73M & 6.82 \\
\bottomrule
\end{tabular}
\label{tab:fps}
\end{table}

\section{Implementation details}
\subsection{Mip-NeRF 360 dataset}
All experiments were conducted using an NVIDIA RTX 4090. Our method was implemented within the Mini-Splatting~\cite{msv1} framework and trained for 30K iterations. At the 20K iteration simplification process, local distinctiveness scoring was incorporated where the factor $\lambda$ was set to 2. The dimension of appearance features $D$ was set to 3.
Scale and rotation were trained from the initial training, while appearance features were introduced at 15K iterations. At this stage, the static features were initialized using the spherical harmonics DC coefficients trained until 15K iterations, whereas view-dependent features were initialized as zero vectors.

From 29K iterations (last 1K iterations), SVQ (Sub-Vector Quantization) was applied to per-Gaussian features. As mentioned in the paper, to enhance training efficiency, K-means clustering was performed once. The assigned indices based on K-means were fixed, and only the codebooks were optimized for the remaining 1K iterations.
For SVQ, different bit allocations were assigned.
\begin{itemize}
\item Scale: length 1, $2^6$ codes for each sub-vector
\item Rotation: length 2, $2^9$ codes for each sub-vector
\item Appearance features: length 2, $2^{10}$ codes for each sub-vector
\end{itemize}
The length 1 SVQ applied to scale can be interpreted as scalar quantization, dynamically learning the quantization range with the codebooks. All codes in the codebook are stored with 16-FP precision.

This SVQ configuration was commonly applied across all variants from XS to XL. The model storage for each variant was determined only by the importance score threshold $\tau$, which is used for simplification at the 20K iteration, set to 0.96, 0.98, 0.99, 0.999, and 0.999, respectively.

\subsection{Zip-NeRF dataset}
We used the 3DGS-MCMC~\cite{mcmc} framework, and tuned it for the Zip-NeRF dataset, by training 3M Gaussians for 150K iterations with a densification interval of 1K iterations. Other hyperparameters are set to default values.
For OMG training, the appearance features were introduced at 100K iterations, and from 140K iterations (last 10K iterations), SVQ was applied to per-Gaussian features.
For SVQ, bit allocations were assigned as follows:
\begin{itemize}
\item Scale: length 1, $2^7$ codes for each sub-vector
\item Rotation: length 2, $2^{10}$ codes for each sub-vector
\item Appearance features: length 2, $2^{11}$ codes for each sub-vector
\end{itemize}

\begin{table}[t]
\centering
\caption{Ablation study on the post-processing methods applied in OMG.}
\begin{tabular}{ccccccc}
\toprule
G-PCC & Huffman & OMG-XS & OMG-S & OMG-M & OMG-L & OMG-XL \\
\cmidrule(lr){1-2}\cmidrule(lr){3-7}
- & -          & 5.82 & 6.83 & 7.66 & 9.47 & 9.89 \\
\checkmark & -   & 4.30 & 5.04 & 5.64 & 6.92 & 7.25 \\
- & \checkmark  & 5.58 & 6.54 & 7.33 & 9.08 & 9.46 \\
\checkmark & \checkmark & \textbf{4.06} & \textbf{4.75} & \textbf{5.31} & \textbf{6.52} & \textbf{6.82} \\
\bottomrule
\end{tabular}
\label{tab:pp}
\end{table}

\section{Effect of post-processings}
As mentioned in the main paper, we applied the following two post-processing methods:
\begin{itemize}
    \item Compressing the 16-bit quantized position with G-PCC~\cite{gpcc}.
    \item Huffman encoding~\cite{huffman} to SVQ indices and compressing the results with LZMA~\cite{lzma}.
\end{itemize}
Both methods are applied losslessly, and we report the resulting storage changes in \Cref{tab:pp}. When applied independently, G-PCC and Huffman encoding consistently reduce the total storage by 26-27\% and 4-5\% across all storage budgets, respectively. Applying both methods together also results in the overall storage reduction remaining consistent at approximately 30-32\%.

\begin{table}[]
\centering
\caption{The average storage allocation for each component across OMG variants. `Actual size' refers to the total size of a single file containing all components. }
\label{tab:breakdown}
\begin{tabular}{lccccc}\toprule
Attribute & OMG-XS & OMG-S & OMG-M & OMG-L & OMG-XL \\\midrule
Position & 0.93 & 1.08 & 1.20 & 1.43 & 1.52 \\
Scale & 0.83 & 0.97 & 1.09 & 1.33 & 1.41 \\
Rotation & 0.87 & 1.02 & 1.15 & 1.40 & 1.49 \\
Appearance & 1.39 & 1.63 & 1.82 & 2.22 & 2.35 \\
MLPs & 0.03 & 0.03 & 0.03 & 0.03 & 0.03 \\\midrule
Total & 4.04 & 4.73 & 5.29 & 6.42 & 6.80 \\\midrule
Actual size & 4.06 & 4.75 & 5.31 & 6.52 & 6.82 \\\bottomrule
\end{tabular}
\end{table}

\section{Storage analysis}
We conducted experiments to analyze the storage requirements of OMG for representing each attribute, as shown in \Cref{tab:breakdown}. Across all variants, OMG allocates approximately 20-25\% of the total storage to position, scale, and rotation, while around 35\% is dedicated to representing appearance attributes, including static and view-dependent color as well as opacity.
The four MLPs for representing local continuity and aggregating appearance attributes exhibit negligible storage requirements, even without extra compression.

\section{Per-scene results}
We report per-scene results in \Cref{tab:perscene360} (Mip-NeRF 360~\cite{mip360}) and \Cref{tab:perscenetntdb} (T\&T~\cite{tnt} and DB~\cite{db}).  

\begin{table*}[]
\centering
\caption{Per-scene results evaluated on the Mip-NeRF 360~\cite{mip360} dataset.}
\label{tab:perscene360}
\resizebox{\textwidth}{!}{
\begin{tabular}{clcccccccccc}\toprule
Method & Metric & bicycle & bonsai & counter & flowers & garden & kitchen & room & stump & treehill & Avg. \\\midrule
\multirow{7}{*}{OMG-XS} & PSNR & 24.95 & 30.90 & 28.40 & 21.32 & 26.42 & 30.81 & 31.09 & 27.00 & 22.60 & 27.06 \\
 & SSIM & 0.743 & 0.932 & 0.899 & 0.596 & 0.818 & 0.919 & 0.918 & 0.788 & 0.647 & 0.807 \\
 & LPIPS & 0.276 & 0.202 & 0.206 & 0.368 & 0.190 & 0.137 & 0.208 & 0.247 & 0.357 & 0.243 \\
 & Train & 18:03 & 20:30 & 24:44 & 19:18 & 18:02 & 23:45 & 20:30 & 17:49 & 19:40 & 20:15 \\
 & \#Gauss & 480772 & 263892 & 310056 & 543034 & 607254 & 356752 & 281236 & 523821 & 479520 & 427371 \\
 & Size & 4.61 & 2.53 & 2.95 & 5.24 & 5.65 & 3.33 & 2.67 & 4.95 & 4.64 & 4.06 \\
 & FPS & 682 & 648 & 433 & 616 & 615 & 498 & 648 & 708 & 658 & 612 \\\cmidrule(lr){1-2}\cmidrule(lr){3-11}\cmidrule(lr){12-12}
\multirow{7}{*}{OMG-S} & PSNR & 25.08 & 31.05 & 28.56 & 21.18 & 26.56 & 30.89 & 31.20 & 27.08 & 22.64 & 27.14 \\
 & SSIM & 0.750 & 0.936 & 0.903 & 0.602 & 0.826 & 0.921 & 0.922 & 0.792 & 0.650 & 0.811 \\
 & LPIPS & 0.264 & 0.195 & 0.199 & 0.358 & 0.177 & 0.132 & 0.201 & 0.239 & 0.347 & 0.235 \\
 & Train & 19:01 & 21:09 & 25:19 & 20:13 & 18:41 & 24:12 & 21:38 & 18:29 & 19:55 & 20:57 \\
 & \#Gauss & 573126 & 310096 & 360930 & 633607 & 691441 & 412126 & 338884 & 619734 & 573425 & 501485 \\
 & Size & 5.46 & 2.94 & 3.41 & 6.10 & 6.43 & 3.83 & 3.19 & 5.83 & 5.54 & 4.75 \\
 & FPS & 601 & 585 & 401 & 555 & 556 & 462 & 620 & 601 & 588 & 552 \\\cmidrule(lr){1-2}\cmidrule(lr){3-11}\cmidrule(lr){12-12}
\multirow{7}{*}{OMG-M} & PSNR & 25.14 & 31.06 & 28.62 & 21.40 & 26.71 & 31.05 & 31.30 & 27.06 & 22.55 & 27.21 \\
 & SSIM & 0.756 & 0.938 & 0.905 & 0.606 & 0.832 & 0.923 & 0.923 & 0.794 & 0.652 & 0.814 \\
 & LPIPS & 0.256 & 0.190 & 0.195 & 0.351 & 0.169 & 0.129 & 0.198 & 0.233 & 0.339 & 0.229 \\
 & Train & 18:58 & 21:01 & 25:44 & 20:35 & 18:51 & 24:18 & 22:14 & 18:31 & 20:22 & 21:10 \\
 & \#Gauss & 646191 & 350999 & 400442 & 708074 & 772338 & 454908 & 375520 & 704907 & 649157 & 562504 \\
 & Size & 6.15 & 3.33 & 3.76 & 6.79 & 7.18 & 4.21 & 3.53 & 6.61 & 6.24 & 5.31 \\
 & FPS & 562 & 536 & 371 & 510 & 522 & 440 & 566 & 566 & 525 & 511 \\\cmidrule(lr){1-2}\cmidrule(lr){3-11}\cmidrule(lr){12-12}
\multirow{7}{*}{OMG-L} & PSNR & 25.24 & 31.47 & 28.66 & 21.45 & 26.83 & 31.03 & 31.26 & 27.05 & 22.57 & 27.28 \\
 & SSIM & 0.762 & 0.941 & 0.907 & 0.613 & 0.837 & 0.924 & 0.926 & 0.795 & 0.653 & 0.818 \\
 & LPIPS & 0.241 & 0.183 & 0.189 & 0.338 & 0.160 & 0.126 & 0.191 & 0.226 & 0.329 & 0.220 \\
 & Train & 19:25 & 21:16 & 26:06 & 20:50 & 19:14 & 24:20 & 22:05 & 19:22 & 21:14 & 21:32 \\
 & \#Gauss & 813561 & 463285 & 480133 & 859963 & 909961 & 524457 & 524457 & 869388 & 819435 & 696071 \\
 & Size & 7.69 & 4.32 & 4.48 & 8.23 & 8.42 & 4.82 & 4.82 & 8.14 & 7.81 & 6.52 \\
 & FPS & 476 & 492 & 332 & 422 & 422 & 405 & 539 & 468 & 414 & 441 \\\cmidrule(lr){1-2}\cmidrule(lr){3-11}\cmidrule(lr){12-12}
\multirow{7}{*}{OMG-XL} & PSNR & 25.22 & 31.51 & 28.78 & 21.52 & 26.93 & 31.15 & 31.25 & 27.00 & 22.69 & 27.34 \\
 & SSIM & 0.764 & 0.942 & 0.908 & 0.614 & 0.839 & 0.925 & 0.926 & 0.796 & 0.655 & 0.819 \\
 & LPIPS & 0.239 & 0.182 & 0.187 & 0.334 & 0.157 & 0.126 & 0.191 & 0.224 & 0.324 & 0.218 \\
 & Train & 20:43 & 21:54 & 26:21 & 22:09 & 20:23 & 24:56 & 22:37 & 20:22 & 22:33 & 22:26 \\
 & \#Gauss & 864124 & 450246 & 507473 & 922061 & 953050 & 547636 & 493754 & 920589 & 885229 & 727129 \\
 & Size & 8.15 & 4.22 & 4.72 & 8.81 & 8.82 & 5.02 & 4.58 & 8.59 & 8.44 & 6.82 \\
 & FPS & 430 & 465 & 324 & 379 & 422 & 397 & 512 & 435 & 384 & 416 \\\bottomrule
\end{tabular}}
\end{table*}

\begin{table*}[]
\centering
\caption{Per-scene results evaluated on the Tank\&Temples~\cite{tnt} and Deep Blending~\cite{db} datasets.}
\label{tab:perscenetntdb}
\resizebox{0.7\textwidth}{!}{
\begin{tabular}{cl|ccc|ccc}\toprule
\multirow{2}{*}{Method} & \multirow{2}{*}{Metric} & \multicolumn{3}{c}{Tank\&Temples} & \multicolumn{3}{c}{Deep Blending} \\\cmidrule(lr){3-5}\cmidrule(lr){6-8}
                        &                         & Train     & Truck     & Avg.      & drjohnson   & Playroom  & Avg.    \\\midrule
\multirow{7}{*}{OMG-M}      & PSNR                    & 21.78     & 25.25     & 23.52     & 29.37       & 30.18     & 29.77   \\
                        & SSIM                    & 0.806     & 0.878     & 0.842     & 0.905       & 0.910     & 0.908   \\
                        & LPIPS                   & 0.233     & 0.144     & 0.189     & 0.253       & 0.253     & 0.253   \\
                        & Train                   & 12:12     & 11:30     & 11:51     & 17:18       & 14:51     & 16:05   \\
                        & \#Gauss                 & 303187    & 257649    & 330418    & 520385      & 404237    & 462311  \\
                        & Size                    & 2.95      & 3.49      & 3.22      & 4.87        & 3.82      & 4.34    \\
                        & FPS                     & 861       & 913       & 887       & 829         & 959       & 894     \\ \cmidrule(lr){1-2}\cmidrule(lr){3-5}\cmidrule(lr){6-8}
\multirow{7}{*}{OMG-L}      & PSNR                    & 21.85     & 25.36     & 23.60     & 29.44       & 30.32     & 29.88   \\
                        & SSIM                    & 0.811     & 0.881     & 0.846     & 0.907       & 0.912     & 0.910   \\
                        & LPIPS                   & 0.225     & 0.136     & 0.181     & 0.247       & 0.247     & 0.247   \\
                        & Train                   & 12:12     & 11:39     & 11:56     & 17:39       & 14:58     & 16:19   \\
                        & \#Gauss                 & 369440    & 442359    & 405900    & 627868      & 485329    & 556599  \\
                        & Size                    & 3.58      & 4.28      & 3.93      & 5.86        & 4.55      & 5.21    \\
                        & FPS                     & 760       & 780       & 770       & 745         & 874       & 810    \\\bottomrule
\end{tabular}}
\end{table*}

\end{document}